\pdfoutput=1
\documentclass[11pt]{article}

\usepackage[preprint]{acl}

\usepackage{times}
\usepackage{latexsym}
\usepackage{float}
\usepackage[T1]{fontenc}
\usepackage[utf8]{inputenc}

\usepackage{booktabs}
\usepackage{multirow}
\usepackage{microtype}
\usepackage{hyperref}
\usepackage{url}
\usepackage{float}
\usepackage{color}
\usepackage{tcolorbox}
\usepackage{CJKutf8}
\usepackage{graphicx} % 用于插入图片
\usepackage{caption} % 可选：用于调整标题格式
\usepackage{colortbl}
\usepackage{xcolor}

\usepackage{CJKutf8}
\usepackage{graphicx} % 用于插入图片
\usepackage{caption} % 可选：用于调整标题格式
\usepackage{inconsolata}
\definecolor{jirailight}{RGB}{227,214,231}    % #e3d6e7 - Light lavender
\definecolor{jiraimedium}{RGB}{204,181,200}   % #ccb5c8 - Muted lilac
\definecolor{jiraipink}{RGB}{182,136,158}     % #b6889e - Dusty rose
\definecolor{jiraimid}{RGB}{130,116,139}      % #82748b - Muted purple
\definecolor{jiraidark}{RGB}{54,45,59}        % #362d3b - Deep aubergine

\usepackage{graphicx}

\title{JiraiBench: A Cross-lingual Benchmark for Evaluating Large Language Models' Detection of Human Risky Health Behavior Content in Jirai Community}

\author{
    \textbf{Yunze Xiao}$^{1\dagger}$\thanks{$^\dagger$Equal Contribution $^\S$Project Leader $^\ddagger $Corresponding Authors },
    \textbf{Tingyu He}$^{2\dagger}$,
    \textbf{Lionel Z. Wang}$^{3\dagger\S
    }$,
    \textbf{Yiming Ma}$^{3}$,\\
    \textbf{Xingyu Song}$^{4}$,
    \textbf{Xiaohang Xu}$^{4}$, 
    \textbf{Mona Diab}$^{1\ddagger}$
    \textbf{Irene Li}$^{4\ddagger}$,
    \textbf{Ka Chung Ng}$^{3\ddagger}$\\
    $^{1}$Carnegie Mellon University $^{2}$University of Washington \\
    $^{3}$The Hong Kong Polytechnic University $^{4}$The University of Tokyo \\
    \texttt{\{yunzex,mdiab\}@cs.cmu.edu},
    \texttt{zhe-leo.wang@connect.polyu.hk},\\
    \texttt{ireneli@ds.itc.u-tokyo.ac.jp},
    \texttt{kc-boris.ng@polyu.edu.hk} \\}
\begin{document}
\begin{CJK*}{UTF8}{min}
\maketitle
\begin{abstract}
In this paper, we present the first cross-lingual dataset that captures a transnational cultural phenomenon, focusing on the Chinese and Japanese "Jirai" subculture and its association with risky health behaviors. Our dataset of more than 15,000 annotated social media posts forms the core of JiraiBench, a benchmark designed to evaluate LLMs on culturally specific content. This unique resource allowed us to uncover an unexpected cross-cultural transfer in which Japanese prompts better handle Chinese content, indicating that cultural context can be more influential than linguistic similarity. Further evidence suggests potential cross-lingual knowledge transfer in fine-tuned models. This work proves the indispensable role of developing culturally informed, cross-lingual datasets for creating effective content moderation tools that can protect vulnerable communities across linguistic borders.
\end{abstract}
{\color{red} \textbf{Disclaimer}: \textit{
This paper describes human content related to risky health behaviors and potentially harmful behaviors that may be disturbing to some readers. }} 
\section{Introduction}
Risky health behaviors (RHB), such as eating disorders (ED), nonsuicidal self-injury (NSSI) and Drug Misuse (DM), constitute a profound public health challenge, often linked to complex mental health conditions \citep{baumeister1988self, van1991childhood, firestone1990suicide}. The digital era has complicated this landscape, as online communities can normalize or reinforce these behaviors, creating new challenges for intervention. What makes these behaviors particularly concerning is their tendency to operate beneath the surface of conventional detection systems, often interwoven with complex mental health conditions such as depression, anxiety, post-traumatic stress disorder, or personality disorders \citep{van1991childhood,firestone1990suicide}.

Previous research on detecting risky health behaviors online has established important methodological foundations but remains limited in real-world applicability. Studies have typically focused narrowly on isolated behaviors like eating disorders (ED) \citep{wang2017detecting, moessner2018analyzing, chancellor2016post}, non-suicidal self-injury (NSSI) \citep{wang2017understanding, un2021towards, ragheb2021negatively}, or drug misuse (DM) \citep{fisher2023automating, nasralah2020social, phan2017enabling, fan2017social}. These approaches address the complex comorbidities common in at-risk individuals. Furthermore, an overwhelming concentration on English content \citep{scherr2020detecting, tebar2021early, sixto2020self} creates significant blind spots in multilingual environments. While community-based detection has shown promise \citep{tebar2021early, wang2017detecting, chancellor2016quantifying}, its limited adoption neglects the crucial, nuanced social dynamics through which these behaviors are normalized. These constraints highlight a collective need for more comprehensive, multilingual, and contextually-aware detection frameworks.

To address these gaps, our study examines the Jirai community (in English: landmine; in Chinese: 地雷; in Japanese: じらい)\footnote{\href{https://aesthetics.fandom.com/wiki/Jirai_Kei}{Jirai Kei, Aesthetics Wiki}}, a transnational phenomenon of social media that spans Chinese and Japanese online spaces. This community is an ideal case study as it simultaneously encompasses multiple RHBs (DM, ED, and NSSI), uses coded language to evade moderation, and operates across different regulatory environments. By analyzing this bilingual community, we demonstrate the necessity of incorporating language-specific nuances and community dynamics into detection systems, addressing the key limitations of existing models.

Our contributions are summarized as follows:

\begin{enumerate}
    \item We conducted comprehensive experiments in four state-of-the-art LLMs using two baseline configurations, systematically evaluating performance in both target languages with prompts in Chinese, Japanese, and English, revealing unexpected patterns of cross-cultural transfer.
    \item We uncover a noteworthy emergent phenomenon in which Japanese instruction prompts consistently outperform Chinese prompts when processing Chinese content, suggesting important linguistic-cultural bridges in transnational content moderation tasks.
    \item We investigate cross-lingual transfer capabilities by fine-tuning Qwen2.5 7B on Chinese data and demonstrate significant performance improvements on both source and target languages, particularly by improving detection of risky health behaviors in Japanese content without explicit Japanese training data.
\end{enumerate}

\begin{figure*}
    \centering
    % Make sure the path 'figure/figure.png' is correct in your project directory
    \includegraphics[width=\textwidth]{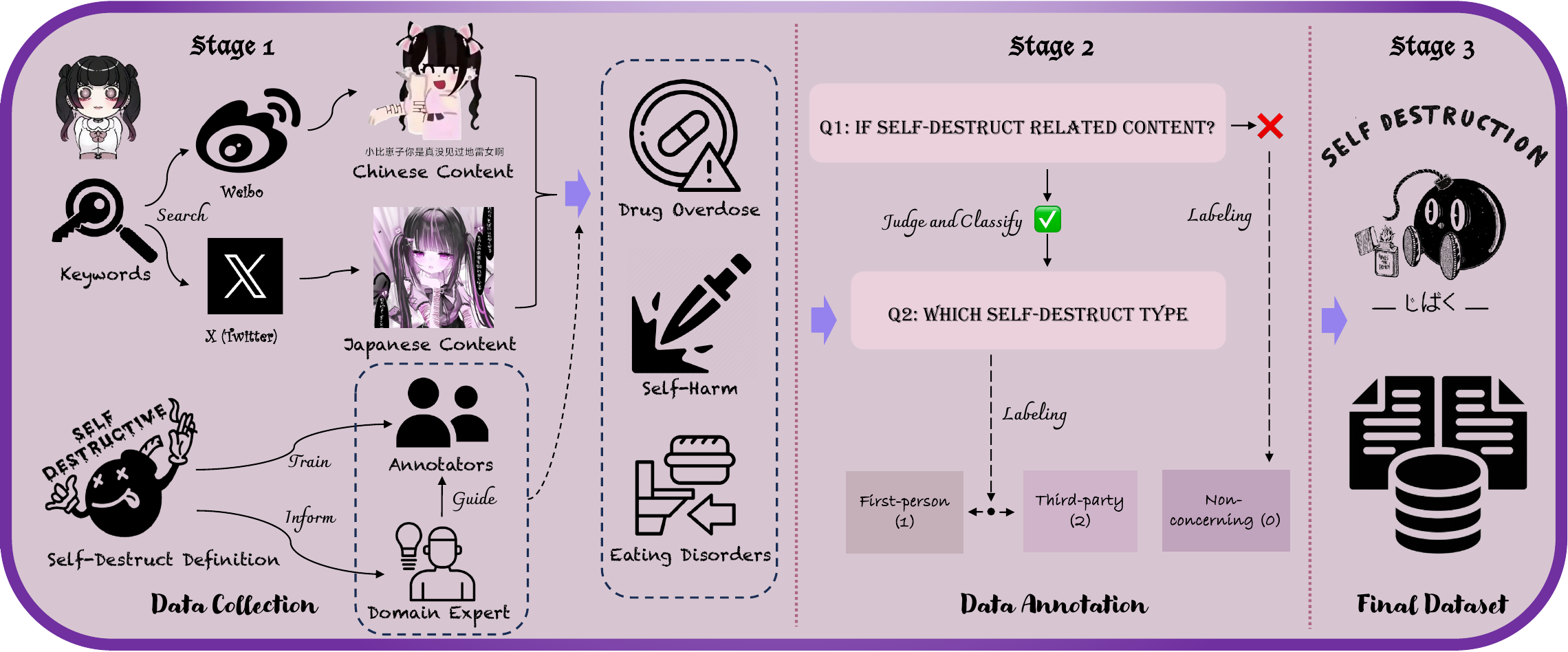} 
    \caption{Illustration of JiraiBench dataset construction procedure. ”0” indicating absence of targeted behaviors, ”1” signifying first-person expressions, and ”2” denoting third-party descriptions.} 
    \label{fig:your_label} 
\end{figure*}
\section{Related Work}
\subsection{Social Media and Mental Health}
Natural language processing (NLP) has long been a pivotal tool for researchers who studied mental health problems through textual data. Social media are one of the valuable resources to understand the mental dynamics of users, with studies showing that online activities of individuals carry detectable signals and can reflect their psychological states \citep{Paul2011, Coppersmith2014}.

Among the considerable number of NLP-based mental health studies, depression is the most discussed because of its high prevalence and broad implications. It has been well established that the mental states of people can be conveyed through language use \citep{Chancellor2020} and that depressed people tend to use first-person pronouns and negative words with a higher frequency \citep{DeChoudhury2013a}. The boundary of NLP research in mental health was also extended to other types as mental illness has become a shared, concerning issue over the society. Post-Traumatic Stress Disorder (PTSD) \citep{murarka2021}, Attention Deficit Hyperactivity Disorder (ADHD) \citep{guntuku2023}, and other general problems have all been examined via language cues, such as references to trauma, heightened expressions of fear,  expressions of worry, and rumination. These linguistic markers are emblematic of negative emotions, exposing critical insights into how mental states manifest in languages \citep{teodorescu-etal-2023-language}.

The evolution of NLP nurtured the growing research interest in this domain. As this field matured in mid-2010s, research focus shifted from domain-specific lexical analysis to machine learning-based predictive models. Behavioral features, such as linguistic style, posting frequency, and social network attributes were engineered and then fed into classifiers, such as logistic regression (LR) and support vector machine (SVM), to detect relevant user-generated content \citep{DeChoudhury2013a, Tsugawa2015}.

The introduction of Transformer-based models revolutionized the research landscape in this area, bringing remarkable improvements in the detection task and outperformed earlier deep learning models \citep{jiang-etal-2020-detection, matero2019suicide}. In particular, specialized models like  MentalBERT that were trained in a mental health-specific corpus have demonstrated an improved capability of identifying edged language patterns in people's mental health disclosures \citep{ji-etal-2022-mentalbert}. 

Although the prior methods have shown promising results, several key challenges persist. One recurring issue is data scarcity and quality. Clinical diagnostic data is normally private and scarce, while datasets populated with online posts were in the doubt of construct validity---the lack of a standardized assessment of each post not only introduces mislabeling risk but also complicates the process of results evaluation. Another pivotal issue lies in the peculiar, intricate linguistic patterns in posts involved with negative mental conditions. Usually, distress was expressed in a figurative way, where creative metaphors, sarcasm, and irony were communicated as an encoded language \citep{Coppersmith2014, mendes-caseli-2024-identifying}. Furthermore, most of the research was developed on an English data set with far fewer in other languages \citep{Cao_2025} and the models therein were trained on a corpus constructed from a single platform or a homogeneous demographic, which consequently limits the applicability of the models to more diverse populations. 

\subsection{Risky Health Behaviors Detections}
Besides the shared attributes, targeting RHBs differs remarkably from the task on general mental illness. RHBs that might cause detrimental and even life-threatening outcomes necessitate more accurate detection to take preventive actions, while their unique linguistic patterns confront researchers with more challenges.

Among these challenges, the use of risky language is a prominent one. Individuals involved in RHBs tend to communicate their situations by using strategically encoded languages---expressions that are linguistically ambiguous, metaphorical, figurative, or culturally unique \citep{yadav-etal-2020-identifying, mendes-caseli-2024-identifying}. Therefore, effective RHBs scrutiny requires systems capable of reasoning over latent implications and culture-specific metaphors.

Flagging users on social media who reveal a tendency towards NSSI and suicidal ideation is a vital yet sensitive application of NLP. Studies reveal that the combination of linguistic signals from content, such as hopelessness, self-hate, and past suicidal attempts, coupled with online behavioral patterns greatly improve the models performance---and it usually gets better when enhanced with domain knowledge \citep{zhang2024ketch, ng2023augmenting,  Choudhury2017TheLO}. Some works focus specifically on NSSI that might include mentions of self-cutting or injury \citep{cliffe2021selfharm}, where the incorporation of psychological lexicons helps models identify depressive expressions as warning sign. 

Likewise, ED detection is another intractable task because the involved individuals barely explicitly state their conditions. Instead, such tendencies manifest themselves through obsessive discussions of weight, food, calories, and body images, making ingestion of specific knowledge essential to ensure the performance of the models in the classification task \citep{chan2022challenges}. 

Social media-based DM study is also one concerning area \citep{WRIGHT2021103824}. One unique challenge is the language ambiguity of drug references and the fast-evolving drug landscape, further complicating the construction of datasets---from identifying emerging drugs through obscure drug references to discerning the subtextual intent. Annotating such data also requires expert assistance in analyzing if the post is truly indicative of DM rather than casual innocuous mentioning.

The emergence of LLM has opened new research avenues, showing strong capabilities in detecting user-generated RHB content and identifying worsening psychological conditions, even in zero-shot or few-shot settings \citep{stade2024}. These models excel at understanding linguistic nuances in intense RHB content, enabling them to detect subtextual signals of mental distress for immediate intervention. However, LLM-based RHB detection faces significant generalizability challenges across platforms, cultures, and languages. Since mainstream LLMs were trained primarily on English-centric corpora with limited representation of other languages, even state-of-the-art models struggle in multilingual environments where community-based, culture-dependent expressions are common. Additionally, while LLMs demonstrate strong contextual understanding, user-specific linguistic peculiarities can compromise performance, particularly when expressions are intentionally obfuscated to evade content moderation \cite{xiao-etal-2024-toxicloakcn}.  

\begin{tcolorbox}[
    colframe=jiraipink, 
    colback=jirailight!80!white,
    coltitle=white,
    fonttitle=\bfseries, 
    title= \small Jirai Kei ,
    boxrule=0.7mm,
    left=5pt, right=5pt, top=5pt, bottom=5pt
]

"Jirai Kei" (in English: landmine style; in Chinese: 地雷系; in Japanese: じらいけい) refers to a transnational social media subculture that emerged in Japan during the early 2020s, gaining particular prominence during pandemic restrictions. The term literally translates to "landmine style", metaphorically representing psychological vulnerabilities that might "explode" when triggered. This community is characterized by individuals who publicly express various forms of psychological distress and self-destructive tendencies across social media platforms. Jirai communities exhibit distinctive communication patterns, including specialized terminology and coded language that conveys emotional vulnerability, particularly related to eating disorders, self-harm behaviors, and substance misuse. This phenomenon spans both Japanese and Chinese online spaces, creating interconnected networks where participants share experiences of psychological distress through culturally specific expressions.
\label{dataexample: kkpuzzle}
\end{tcolorbox}

\section{Dataset Construction}
In this section, we describe the construction of the JiraiBench dataset. We start with introducing the data collection and filtering process, then turning to the annotation.

\subsection{Data Collection and Cleaning}
For the Chinese dataset, we implemented a keyword-based search approach to gather relevant posts from Sina Weibo\footnote{\url{https://weibo.com/}}, China's equivalent of X (formerly Twitter). Using specialized lexicons developed by community experts in both languages for DM, ED, and NSSI, we identified posts that discuss RHBs within the community. This process yielded a comprehensive Chinese dataset containing 10,419 potentially concerning posts. The lexicons were meticulously curated to include both explicit and implicit terminology commonly associated with each type of behavior. 

Similarly, we applied the same methodological framework to construct the Japanese dataset, ensuring consistency between the collections. From X\footnote{\url{https://x.com/}}, we collected 5,000 tweets, with varying associations to the three behaviors of interest. Both datasets underwent an identical data cleansing pipeline to remove irrelevant content, including advertisements, duplicate entries, excessively short posts lacking meaningful information, and other noise. In accordance with the platform policies and to maintain user privacy, we eliminated all usernames, gender information, and other personally identifiable information. We present randomly selected data examples in Appendix \ref{sec:example}.

\subsection{Data Annotation}
Our annotation protocol employs a multidimensional framework to identify RHBs in social media texts, establishing three distinct categories: Drug Poisoning (DM), Eating Disorders (ED), and Non-suicidal Self-Injury (NSSI). This breakdown enables fine-grained, category-based analysis. For each post, annotators evaluated the content using a three-point ordinal scale: "0" for the absence of targeted behaviors, "1" for first-person expressions (e.g. "I relapsed again"), and "2" for third-party descriptions or general discussions (e.g. "It is dangerous to mix these pills"), a distinction that provides critical context for analysis and intervention.

The annotation process followed a structured, multi-step methodology. Instead of relying on general impressions, each annotator was required to perform a complete \textbf{reading}, which meant analyzing the full textual content of the post, including paralinguistic markers such as emojis and relevant hashtags that could indicate tone or intent. Following this initial reading, they applied a set of \textbf{precise annotation criteria}, which were developed prior to the main annotation phase to ensure consistency. These criteria, detailed in Appendix \ref{sec:annoGuide}, provided explicit definitions for each category:
\begin{itemize}
    \item \textbf{DM} was defined as any clear mention of substance or drug misuse or dangerous levels of drug consumption.
    \item \textbf{ED} was defined as content capturing extreme food restrictions, binge-purge behaviors, or an unhealthy fixation on weight.
    \item \textbf{NSSI} was defined as discussions of suicidal ideation or direct mentions of self-injury behaviors.
\end{itemize}
Annotators were specifically trained with these guidelines to distinguish genuine expressions of harmful tendencies from figurative or metaphorical language. Each of the three dimensions (DM, ED, NSSI) was evaluated independently for every post, allowing for multi-label classification where a single post might contain evidence of multiple behaviors.

For this research, we recruited a team of six annotators, evenly divided between Chinese and Japanese, and spent over 100 hours training and labeling the annotators on each side. For each language, the annotation team comprised three members: two native speakers and one community expert, applied to both Chinese and Japanese. All annotators received equitable compensation at minimum wage levels applicable in their respective regions (Hong Kong SAR for Chinese and Japan for Japanese). Annotator disagreements were systematically resolved through a combination of majority voting and expert supervision, ensuring annotation consistency. We evaluated inter-annotator reliability using both pairwise and overall agreement metrics shown in Appendix \ref{sec:IAA}. The results demonstrate substantial agreement across all annotation dimensions, confirming the reliability of our annotation framework for the detection of RHBs. Specific annotation guidelines can be found in the Appendix \ref{sec:annoGuide}.

\subsection{Data Composition}
Our analysis includes two datasets: 10,419 posts from the Chinese social platform Sina Weibo and 5,000 posts from Japanese X (formerly Twitter). The Chinese dataset shows higher content related rates in all categories, with DM-related posts that contain 30.55\% first-person expressions compared to only 3.82\% in the Japanese dataset. Similarly, the content of ED in the Chinese data set showed 15.04\% first-person expressions versus 3.14\% in the Japanese dataset. NSSI content demonstrated more comparable distributions, with first-person expressions comprising 10.27\% of Chinese posts and 8.36\% of Japanese posts. Table \ref{sec:data_dist} provides a comprehensive breakdown of both datasets across all behavior categories and label types.

\section{Experiments}
Our experimental framework systematically evaluates the performance of LLMs in detecting risky health behaviors in social media content in multiple languages and test conditions. We employ four state-of-the-art language models, Llama-3.1 8B \citep{llama3}, Qwen-2.5 7B \citep{qwen2.5}, DeepSeek-v3 \citep{deepseekv3}, as well as our finetuned JiraiLLM-Qwen. We then structure our investigations around three complementary experimental paradigms.

In our initial experimental design, we attempted to include GPT-4o as a baseline; however, the model consistently refused to perform classification tasks that involve the content of the RHB despite various prompting strategies. This limitation highlights significant limitations in the evaluation of closed source models for the detection of sensitive content and underscores the need for specialized research interfaces that balance safety mechanisms with legitimate research on harmful content detection.
\subsection{Baseline}
The baseline experiments establish fundamental performance benchmarks across both Chinese and Japanese corpora under zero-shot and few-shot learning conditions. For zero-shot evaluation, models receive task instructions without exemplars, requiring them to leverage pre-trained knowledge for classification across our three-dimensional annotation scheme. The few-shot configuration provides two examples that represent diverse manifestations of risky health behaviors with balanced representation across categories. To avoid selective bias, all examples shown in the data are randomly sampled from our dataset. We independently calculated precision, recall, and F1 scores for each dimension to assess overall classification effectiveness and identify language-specific processing disparities. The prompts are provided in the appendix \ref{sec:prompt}.

\subsection{Crosslingual Transfer}
The evaluation of the effectiveness of cross-lingual transfer in our study involved fine-tuning Qwen2.5 with 3,000 randomly sampled data points from a Chinese data set utilizing Chinese prompts, resulting in the development of JiraiLLM-Qwen. Our comprehensive evaluation framework assessed the model's performance across both the source Chinese dataset and the target Japanese dataset, providing empirical insights into the model's capacity to transfer linguistic knowledge between related but distinct language systems without explicit training on the target language. 

This investigation offers valuable insight into the capabilities of cross-linguistic generalization in LLMs, particularly in East Asian language contexts where shared writing systems and linguistic features may facilitate knowledge transfer despite significant structural and lexical differences between languages. Detailed training parameters and hyperparameter configurations are attached in the appendix \ref{sec:param}.

\subsection{Evaluation Metric}
To align with the established research norm \citep{yang-etal-2024-identifying}, we use the Macro F1 score as the evaluation metrics for our risky health behavior detection task. The metric assesses the model's performance in successfully identifying risky health behaviors.

\begin{table*}[ht]
\centering
\tiny
\resizebox{\textwidth}{!}{%
\begin{tabular}{llllcccccc}
\toprule
\textbf{Dataset} & \textbf{Prompt} & \textbf{Method} & \textbf{Task} 
& \textbf{Qwen2.5 7B} & \textbf{Llama3.1 8B} & \textbf{DeepSeek-v3} 
& \textbf{Jirai-Qwen (CN)} & \textbf{Jirai-Qwen (JP)} & \textbf{Random} \\
\midrule
\multirow{18}{*}{Chinese} 
 & \multirow{6}{*}{Chinese} & \multirow{3}{*}{zero-shot} 
 & DM   & 0.5052 & 0.3598 & 0.5478 & \cellcolor{red!20}{0.692}& 0.6914 & 0.2504 \\
 &  &  & ED   & 0.4706 & 0.3349 & 0.4163 & \cellcolor{red!20}{0.6503}& 0.6997 & 0.3015 \\
 &  &  & NSSI & 0.3927 & 0.3232 & 0.4205 & \cellcolor{red!20}{0.573}& 0.5714 & 0.3076 \\
 &  & \multirow{3}{*}{two-shot} 
 & DM   & 0.4190 & 0.4400 & - & - & - & 0.2504 \\
 &  &  & ED   & 0.2951 & 0.3590 & - & - & - & 0.3015 \\
 &  &  & NSSI & 0.4284 & 0.3714 & - & - & - & 0.3076 \\
\cmidrule(lr){2-10}
 & \multirow{6}{*}{Japanese} & \multirow{3}{*}{zero-shot} 
 & DM   & \cellcolor{cyan!20}{0.5226} & \cellcolor{cyan!20}{0.3969} & 0.5229 & 0.6371 & 0.6557 & 0.2504 \\
 &  &  & ED   & \cellcolor{cyan!20}{0.5431} & \cellcolor{cyan!20}{0.3403} & 0.3962 & 0.6069 & 0.6685 & 0.3015 \\
 &  &  & NSSI & \cellcolor{cyan!20}{0.4154} & \cellcolor{cyan!20}{0.3240} & 0.3670 & 0.5402 & 0.5761 & 0.3076 \\
 &  & \multirow{3}{*}{two-shot} 
 & DM   & 0.3078 & 0.2503 & - & - & - & 0.2504 \\
 &  &  & ED   & 0.3470 & 0.3014 & - & - & - & 0.3015 \\
 &  &  & NSSI & 0.3118 & 0.3081 & - & - & - & 0.3076 \\
\cmidrule(lr){2-10}
 & \multirow{6}{*}{English} & \multirow{3}{*}{zero-shot} 
 & DM   & 0.4796 & 0.3516 & 0.5087 & 0.4769 & 0.5600 & 0.2504 \\
 &  &  & ED   & 0.4023 & 0.2771 & 0.4148 & 0.6173 & 0.5912 & 0.3015 \\
 &  &  & NSSI & 0.3933 & 0.3164 & 0.4286 & 0.5993 & 0.5704 & 0.3076 \\
 &  & \multirow{3}{*}{two-shot} 
 & DM   & 0.3304 & 0.3352 & - & - & - & 0.2504 \\
 &  &  & ED   & 0.2223 & 0.2173 & - & - & - & 0.3015 \\
 &  &  & NSSI & 0.4278 & 0.2773 & - & - & - & 0.3076 \\
\midrule
\multirow{18}{*}{Japanese} 
 & \multirow{6}{*}{Chinese} & \multirow{3}{*}{zero-shot} 
 & DM   & 0.3813 & 0.2869 & 0.5521 & 0.3167 & 0.5007 & 0.3232 \\
 &  &  & ED   & 0.4139 & 0.2573 & 0.4097 & 0.4163 & 0.4514 & 0.3184 \\
 &  &  & NSSI & 0.4241 & 0.3280 & 0.4159 & 0.3114 & 0.3600 & 0.3131 \\
 &  & \multirow{3}{*}{two-shot} 
 & DM   & 0.4087 & 0.3254 & - & - & - & 0.3232 \\
 &  &  & ED   & 0.4172 & 0.2459 & - & - & - & 0.3184 \\
 &  &  & NSSI & 0.4319 & 0.3448 & - & - & - & 0.3131 \\
\cmidrule(lr){2-10}
 & \multirow{6}{*}{Japanese} & \multirow{3}{*}{zero-shot} 
 & DM   & 0.4059 & 0.2430 & \cellcolor{green!20}{0.5974} & 0.5117 & 0.4366 & 0.3232 \\
 &  &  & ED   & 0.4665 & 0.2000 & 0.4585 & \cellcolor{green!20}{0.5761} & 0.4491 & 0.3184 \\
 &  &  & NSSI & 0.4292 & 0.2255 & 0.4782 & \cellcolor{green!20}{0.4840} & 0.3800 & 0.3131 \\
 &  & \multirow{3}{*}{two-shot} 
 & DM   & 0.3428 & 0.3232 & - & - & - & 0.3232 \\
 &  &  & ED   & 0.3684 & 0.3184 & - & - & - & 0.3184 \\
 &  &  & NSSI & 0.3131 & 0.3130 & - & - & - & 0.3131 \\
\cmidrule(lr){2-10}
 & \multirow{6}{*}{English} & \multirow{3}{*}{zero-shot} 
 & DM   & 0.3683 & 0.2385 & 0.5335 & 0.4939 & 0.4106 & 0.3232 \\
 &  &  & ED   & 0.3777 & 0.2661 & 0.4282 & 0.5328 & 0.4151 & 0.3184 \\
 &  &  & NSSI & 0.4302 & 0.3028 & 0.4198 & 0.4738 & 0.3844 & 0.3131 \\
 &  & \multirow{3}{*}{two-shot} 
 & DM   & 0.4210 & 0.2141 & - & - & - & 0.3232 \\
 &  &  & ED   & 0.3940 & 0.2109 & - & - & - & 0.3184 \\
 &  &  & NSSI & 0.4648 & 0.2795 & - & - & - & 0.3131 \\
\bottomrule
\end{tabular}
}
\caption{Performance comparison across models, datasets, prompts, and methods, with tasks organized by rows. \textcolor{red}{Red} values indicate the best approach in Chinese tasks; 
\textcolor{green}{Green} values indicate the best approach in Japanese tasks; 
and \textcolor{cyan}{Cyan} values highlight emergent cross-cultural transfer behavior, which are statistically significantly better than their Chinese zero-shot counterparts.}
\label{tab:performance}
\end{table*}

\section{Result and Discussion}
Table \ref{tab:performance} presents the RHB detection outcomes for all models, showing that Jirai-qwen achieves the highest performance on Chinese data across all three detection dimensions. However, both open-source models exhibit a remarkable pattern in which Japanese instruction prompts consistently outperform Chinese prompts when processing Chinese content, despite Japanese not being the native language of the content. This indicates a significant cross-cultural transfer effect in RHB content detection.

\subsection{Effects on Different Instruction Language}
Our experimental findings reveal compelling emergent cross-cultural transfer patterns that challenge conventional assumptions about language-model alignment in content moderation tasks. In particular, we observe that Japanese instruction prompts often outperform Chinese prompts when moderating Chinese content, particularly in zero-shot scenarios. This surprising pattern suggests a linguistic-cultural bridge effect that transcends simple language matching. The phenomenon manifests itself across multiple model architectures, pointing toward a systematic cross-cultural transfer mechanism rather than an artifact of any single model \cite{wendler-etal-2024-llamas}.

We hypothesize that the effectiveness of Japanese instructions stems from the deep cultural relationship between Japanese and Chinese-speaking communities, particularly in the domain of RHBs, where the "Jirai" cultural framework originated in Japan before being distributed in Chinese-speaking contexts. Japanese prompts may more effectively activate nuanced cultural schemata associated with RHB discourse, directing model attention toward subtle linguistic markers that Chinese instructions, despite their native language advantage, fail to adequately emphasize.

However, this cross-cultural prompt effect shows significant architecture sensitivity, with Japanese prompts producing inconsistent gains across different LLMs. Models like Qwen2.5 7B can incorporate region-specific pretraining data or tokenization schemes that favor prompt-content language alignment. Safety tuning and censorship mechanisms emerge as critical confounding factors, suppressing culturally sensitive tokens differentially across models and impacting generalization. This was particularly evident in the GPT-4o trials, which frequently refused classification tasks involving sensitive content, demonstrating how alignment and safety tuning can inadvertently impede legitimate content moderation research.

Our findings resonate with \citeauthor{wendler-etal-2024-llamas}'s mechanistic interpretability work, which identified a three-phase processing pattern in Llama-2 models handling multilingual input. Their analysis shows that while middle-layer embeddings successfully encode relevant concepts, they systematically favor English tokens even under non-English prompting, revealing that LLMs navigate through an inherently English-biased conceptual space. This bias paradoxically facilitates effective transfer between culturally proximate languages like Japanese and Chinese, where shared cultural substrates (logographic writing systems, historical conceptual exchange) establish pathways for detecting semantically complex phenomena like RHBs more robustly when instruction languages resonate culturally rather than linguistically with target content. These insights have profound implications for multilingual AI in mental health screening and content moderation, suggesting that optimal instruction language selection depends less on linguistic correspondence and more on cultural schema activation. Strategic instruction selection based on shared cultural histories and conceptual frameworks can unlock superior performance, particularly for culturally embedded and psychologically nuanced content.
\begin{table*}[t]
\centering
\small
\setlength{\tabcolsep}{5pt}
\begin{tabular}{lllrrrr}
\toprule
Model (train) & Eval & Prompt & OD & ED & SH & Avg \\
\midrule
\multirow{6}{*}{Jirai-Qwen-cn-LoRA}& \multirow{3}{*}{CN} & Chinese  & 0.6082 & 0.6000 & 0.5678 & 0.5920 \\
&                        & Japanese & 0.6622 & 0.6532 & 0.5743 & 0.6299 \\
&                        & English  & 0.5476 & 0.5624 & 0.5382 & 0.5494 \\
\cmidrule(l){2-7}
& \multirow{3}{*}{JP} & Chinese  & 0.5071 & 0.4836 & 0.3845 & 0.4584 \\
&                      & Japanese & 0.4647 & 0.5120 & 0.3980 & 0.4582 \\
&                      & English  & 0.4220 & 0.4109 & 0.4062 & 0.4131 \\
\midrule
\multirow{6}{*}{Jirai-Qwen-jp-LoRA}& \multirow{3}{*}{CN} & Chinese  & 0.6089 & 0.6001 & 0.5671 & 0.5920 \\
&                        & Japanese & 0.6598 & 0.6535 & 0.5735 & 0.6289 \\
&                        & English  & 0.5475 & 0.5639 & 0.5392 & 0.5502 \\
\cmidrule(l){2-7}
& \multirow{3}{*}{JP} & Chinese  & 0.5068 & 0.4849 & 0.3846 & 0.4587 \\
&                      & Japanese & 0.4721 & 0.5100 & 0.3962 & 0.4595 \\
&                      & English  & 0.4218 & 0.4119 & 0.4064 & 0.4134 \\
\bottomrule
\end{tabular}
\caption{F1 scores by evaluation language, prompt language, and task for two Qwen2.5-7B LoRA variants. Values are shown to 4 decimals.}
\label{tab:qwen_lora_f1_compact}
\end{table*}

\subsection{Effects on Zero-shot VS Few-shot prompting}
The contrasting performance patterns between zero-shot and few-shot methodologies in our experimental framework reveal nuanced dynamics in the detection of cross-cultural risky health behaviors. Our analysis demonstrates that zero-shot approaches consistently outperform few-shot methods across most model language configurations, contradicting conventional wisdom regarding few-shot learning benefits. This phenomenon likely stems from the culturally embedded nature of risky health behaviors, which manifest through subtle linguistic and contextual patterns that resist straightforward exemplification. The cultural scheme underlying RHB in East Asian contexts, particularly within Jirai cultural frameworks originating from Japan, involve deeply situated knowledge that few-shot exemplars struggle to adequately represent. These cultural understandings encompass complex sociolinguistic signals, indirect expressions, and culturally specific metaphors that may be entirely absent from model training distributions.

Performance degradation observed in few-shot settings potentially indicates interference effects, whereby provided examples inadvertently constrain model attention to narrow manifestations of concerning behavioral patterns that fail to generalize across the diverse expression patterns present in authentic data \citep{yang-etal-2022-seqzero}. This limitation appears particularly pronounced when the few-shot examples lack the cultural depth necessary to activate appropriate interpretive frameworks. The complex interrelationship between instruction language and exemplar effectiveness further compounds these challenges, suggesting that the cultural alignment between prompt language and dataset requires more sophisticated calibration than few-shot learning easily accommodates.

\subsection{Cultural vs. Orthographic Transfer}
To address the question of whether the advantages of Japanese prompts come from cultural framing or mere script overlap, we conducted an ablation experiment where all Kanji characters were stripped from Japanese prompts and replaced with phonetic kana. This manipulation allows us to isolate the contribution of shared logographic elements while preserving cultural semantic framing. The model's performance remained relatively strong despite the increased difficulty due to the loss of semantic precision in Kanji-to-Hiragana mapping.

Table \ref{tab:ablation} presents the Macro-F1 scores for both the base model (zero-shot) and fine-tuned model across the three detection dimensions. All results are statistically significant using paired t-test ($p < 0.05$).

\begin{table}[h]
\centering

\resizebox{0.5\textwidth}{!}{%
\begin{tabular}{lccc}
\toprule
Task & Base Model & Fine-tuned Model & $\Delta$F1 \\
\midrule
DM & 0.4003 & 0.4720 & +0.0717 \\
ED & 0.4220 & 0.4930 & +0.0710 \\
NSSI & 0.3587 & 0.3732 & +0.0145 \\
\bottomrule
\end{tabular}
}
\caption{Ablation results showing performance with Kanji-stripped Japanese prompts}
\label{tab:ablation}
\end{table}

This result supports our hypothesis that cross-lingual transfer is driven by cultural framing rather than mere script overlap, as performance remains stable even when shared characters are removed. The continued advantage of Japanese prompts,even in a reduced-script setting,demonstrates that semantic framing rooted in the Jirai cultural context plays a larger role than orthographic similarity.

\subsection{Fine-tuning for Cross-lingual Transfer}
\label{sec:ft_transfer}

We study whether parameter-efficient fine-tuning improves risky health behavior detection and whether gains transfer across languages, and we compare these findings to our prior fine-tuning experiment (Table~\ref{tab:performance}). Starting from Qwen2.5-7B-Instruct, we train two LoRA variants using the same recipe, \texttt{JiraiQwen-CN-LoRA} and \texttt{JiraiQwen-JP-LoRA}. In our main run, we fine-tune on 3,000 randomly sampled Chinese posts with supervised labels under the Chinese prompt template for 3 epochs on 4 NVIDIA A6000 GPUs (per-device batch size 5, gradient accumulation 10, effective batch size 50), keeping other hyperparameters at default values. We then evaluate on both Chinese and Japanese test sets while varying instruction language (Chinese, Japanese, English) to quantify prompt effects.

Compared to the results of the base model in Table~\ref{tab:performance}, supervised adaptation again produces large improvements in the domain in Chinese. Under Chinese prompts on the Chinese dataset, Qwen2.5 increases from an average Macro-F1 of 0.456 (DM, ED, NSSI) to 0.638 with \texttt{JiraiQwen} (CN), confirming that fine-tuning substantially strengthens detection in the source domain. The \texttt{JiraiQwen-\{CN,JP\}-LoRA} experiment isolates two factors not previously disentangled. 

First, the LoRA variant label has negligible impact: across all evaluation settings, \texttt{JiraiQwen-CN-LoRA} and \texttt{JiraiQwen-JP-LoRA} differ by at most 0.0012 Macro-F1, suggesting that the observed behaviors are not driven by which LoRA variant is used under our recipe. 

Second, the effects of the instruction language are strong and asymmetric. In the Chinese test set, the Japanese prompts remain best after fine-tuning (0.629 Macro-F1) compared to Chinese (0.592) and English (0.550), reproducing the advantage of the cross-cultural prompt under controlled adaptation. On the Japanese test set, overall performance remains substantially lower (0.444 averaged across prompts and LoRA variants), Chinese and Japanese prompts are effectively tied (both around 0.459), and English prompts are consistently worse (0.413). 

Together, these results show that fine-tuning yields strong in-domain performance, and that instruction language can materially shape what the model attends to at test time. In particular, for Chinese evaluation, Japanese prompts consistently achieve the highest F1 across both CN-LoRA and JP-LoRA, even outperforming Chinese prompts, which is consistent with our hypothesis that Japanese instructions more effectively activate the Jirai-origin cultural schema that structures RHB discourse and later diffused into Chinese-speaking contexts. s.

\section{Conclusion}
JiraiBench represents the first cross-lingual benchmark for evaluating LLMs' capability to detect risky health behaviors across Chinese and Japanese online communities. Our comprehensive experiments across four state-of-the-art models reveal significant limitations in current systems' effectiveness in identifying RHB content within these linguistically and culturally complex environments. The observed emergent pattern, where Japanese instruction prompts consistently outperform Chinese counterparts, underscores critical cultural-linguistic alignment effects stemming from the Jirai phenomenon's Japanese origins. This counterintuitive discovery demonstrates that cultural proximity can sometimes outweigh linguistic similarity in cross-lingual detection tasks.

Future work should focus on developing more robust cross-cultural transfer learning methodologies, expanding benchmark datasets to include additional languages and cultural contexts, and incorporating more nuanced annotation schemes that capture the complex manifestations of RHBs across diverse communities.

\section*{Limitations and Ethical Statement}
This research engages with the complex ethical considerations involved in detecting risky health behaviors within multilingual online communities. Given the sensitive nature of the content, such as discussions of drug misuse, eating disorders, and nonsuicidal self-injury, we recognized the importance of ethical oversight. The IRB office at the main institution have confirmed in an official communication that social media research of this nature does not qualify as human subjects research and therefore does not require IRB review or exemption. Accordingly, IRB review was not sought. Our study follows established ethical standards for working with publicly available online data.

We recognize valid concerns about the sensitive nature of these data and the vulnerability of the individuals whose content comprises our data set. The decision to conduct this research was not taken lightly. We believe that the technical contributions, specifically the development of culturally informed detection systems that can identify at-risk individuals across linguistic boundaries, provide significant potential benefits that justify the careful use of these data. Enhanced detection capabilities can enable earlier intervention and support for vulnerable individuals in online communities where traditional mental health resources may be inaccessible.

To address privacy concerns, we implemented comprehensive data anonymization protocols that exceed standard practices.
\begin{itemize}
    \item Complete removal of all personally identifiable information (PII), including usernames, user IDs, timestamps, location data, and platform-specific identifiers
    \item Systematic review of content to redact any self-disclosed personal information embedded within posts
    \item Restriction of raw data access to verified academic institutions with demonstrated ethical review processes
\end{itemize}

To ensure responsible data handling across our multiinstitutional collaboration, we implemented strict protocols governing author access to sensitive data. All data annotation activities were conducted exclusively on datasets from which personally identifiable information (PII) had been thoroughly removed prior to any collaborative work. Authors affiliated with institutions other than [main institution] worked solely with these anonymized datasets, with several coauthors serving as community experts who contributed specialized knowledge to the annotation process after PII removal. These collaborating authors, including community experts, provided valuable contributions through annotation of deidentified content, model design, methodological discussions, and manuscript preparation without ever accessing raw data containing personal information. 

Throughout the research process, all coauthors from external institutions had no direct contact with private or identifiable information, instead working with sanitized datasets while contributing feedback on analytical approaches, sharing code for computational analysis, and providing domain expertise in annotation tasks. This collaborative structure allowed us to benefit from diverse expertise, including specialized community knowledge essential for accurate annotation, while maintaining complete separation from original sensitive datasets and upholding the highest standards of data protection and ethical responsibility in handling sensitive health-related online content.

We acknowledge that despite these measures, the absence of explicit consent from content creators remains an ethical limitation. However, we believe that the potential harm from not developing effective detection systems, which could result in missed opportunities for intervention, outweighs the carefully mitigated risks of our approach. Our controlled data release policy ensures that access is limited to researchers who demonstrate both technical competence and ethical commitment to protecting vulnerable populations.

Our annotation protocols prioritized the well-being of our research team. The six annotators received comprehensive training on the recognition of harmful content patterns while maintaining an appropriate emotional distance, with access to mental health resources throughout the annotation process. Compensation substantially exceeded minimum wage standards in their respective regions, reflecting both the specialized expertise required and the emotionally intensive nature of the work.

We remain committed to the responsible development of AI systems that can identify content patterns concerning while respecting the principles of privacy, cultural sensitivity, and human dignity. Future iterations of this work will explore alternative methodologies that can further minimize privacy risks while maintaining detection effectiveness, including synthetic data generation and federated learning approaches that eliminate the need for centralized data collection.

\section*{Acknowledgment}

This work was supported by JST CREST Grant Number JPMJCR21M2. This work was also supported by JST ACT-X (Grant JPMJAX24CU) and JSPS KAKENHI (Grant 24K20832).

\bibliography{custom,latex/anthology_0,latex/anthology_1}

\appendix

\section{Data Distribution}
\label{sec:data_dist}
We present the JiraiBench data distribution in Table \ref{tab:dataset_comparison}.
\begin{table*}[ht]
\centering
\resizebox{\textwidth}{!}{%
\begin{tabular}{llrrrr}
\toprule
\multirow{2}{*}{\textbf{Behavior Category}} & \multirow{2}{*}{\textbf{Label Type}} & \multicolumn{2}{c}{\textbf{Chinese Dataset}} & \multicolumn{2}{c}{\textbf{Japanese Dataset}} \\
\cmidrule(lr){3-4} \cmidrule(lr){5-6}
 &  & \textbf{Count} & \textbf{Percentage} & \textbf{Count} & \textbf{Percentage} \\
\midrule
\multirow{3}{*}{\textbf{Drug Misuse (DM)}} & Non-concerning (0) & 6,268 & 60.16\% & 4,706 & 94.12\% \\
 & First-person (1) & 3,183 & 30.55\% & 191 & 3.82\% \\
 & Third-party (2) & 968 & 9.29\% & 103 & 2.06\% \\
\midrule
\multirow{3}{*}{\textbf{Eating disorders (ED)}} & Non-concerning (0) & 8,605 & 82.59\% & 4,572 & 91.44\% \\
 & First-person (1) & 1,567 & 15.04\% & 157 & 3.14\% \\
 & Third-party (2) & 247 & 2.37\% & 271 & 5.42\% \\
\midrule
\multirow{3}{*}{\textbf{Non-Suicidal Self-Injury (NSSI)}} & Non-concerning (0) & 8,924 & 85.65\% & 4,427 & 88.54\% \\
 & First-person (1) & 1,070 & 10.27\% & 418 & 8.36\% \\
 & Third-party (2) & 425 & 4.08\% & 155 & 3.10\% \\
\midrule
\textbf{Total Posts} & - & 10,419 & 100\% & 5,000 & 100\% \\
\bottomrule
\end{tabular}%
}
\caption{Comparison of Chinese and Japanese Datasets for Concerning Behaviors}
\label{tab:dataset_comparison}
\end{table*}
\section{Data Example}
\label{sec:example}
We present randomly selected data examples in Table \ref{tab:examples}.
\begin{table*}[htbp]
\centering
\resizebox{\textwidth}{!}{%
\begin{tabular}{p{1.5cm}p{6cm}p{6cm}}
\toprule
\textbf{Category} & \textbf{Chinese Example} & \textbf{Japanese Example} \\
\midrule
Drug Misuse (DM) & \CJKfamily{gbsn}喜欢ymsf的濒死感。每次o都想写遗书给朋友留点谷。\textit{(I like the near-death feeling from taking Dextromethorphan. Every time I overdose, I want to write a suicide note and leave some final words to my friends.)} & \CJKfamily{min}ODもアムカも、やってみようかなくらいの気持ちだわ。 \textit{(I feel like I might try an overdose on amoxicillin (Amca), just considering it a bit)} \\
\midrule
Eating Disorder (ED) & \CJKfamily{gbsn}暴食后的催吐让人恶心，父母的控制欲令我窒息。 \textit{(Purging after binge eating makes me sick, and my over-controlling parents suffocate me.)} & \CJKfamily{min}昨日から食べても食べても吐いてしまうくぽ… \textit{(Since yesterday, no matter how much I eat, I keep throwing up...")} \\
\midrule
Non-sucidal self injury (NSSI) & \CJKfamily{gbsn}每次划开手腕的时候，那种痛感反而让我感到平静。 \textit{(Every time I cut my wrist, the pain actually makes me feel calm.)} & \CJKfamily{min}誰も気づかないように、太ももを切った。痛みだけが本当の感情。 \textit{(I cut my thigh where no one would notice. The pain is the only real feeling.)} \\
\bottomrule
\end{tabular}
}
\caption{Paraphrased example of Risky health behavior Content Categories in Chinese and Japanese}
\label{tab:examples}
\end{table*}

\section{Annotation Guidelines}
\label{sec:annoGuide}
\subsection{Annotation Objectives and Framework}
We conducted independent annotation for each social media post across three dimensions:

\begin{enumerate}
    \item \textbf{Drug Misuse (DM)}: Content containing risky health behavior related to drug misuse or extremely dangerous behaviors.
    \item \textbf{Eating Disorders (ED)}: Content that contains eating disorder tendencies, such as anorexia, binge eating, or purging behaviors.
    \item \textbf{Non-Suicidal Self-Injury and Suicidal Ideation (NSSI/SI)}: Content containing nonsuicidal self-injury, suicidal thoughts, or suicide plans.
\end{enumerate}

For each category (DM, ED, NSSI/SI), annotation values were assigned as follows:
\begin{itemize}
    \item \textbf{0}: No indication of the targeted behavior in the post
    \item \textbf{1}: Post implies/reveals the poster's own tendencies in this category (or potential tendencies)
    \item \textbf{2}: Third-party narratives or comments about this category (not the poster's own behavior or tendencies)
\end{itemize}

Note: If a post received a value of 0 across all three categories (DM, ED, NSSI/SI), it was still retained in the dataset (indicating the post did not manifest any risky health behavior-related content).

\subsection{Annotation Procedure}

\subsubsection{Complete Reading}
\begin{itemize}
    \item Read the entire post content, including text, emoticons/emojis, or punctuation that might affect semantics.
    \item For lengthy posts, focus on keywords or contexts related to drugs, diet, bodily harm/suicide, etc.
\end{itemize}

\subsubsection{Identification and Classification}
\begin{itemize}
    \item Determine whether the post relates to any of the following categories: DM, ED, NSSI/SI.
    \item If multiple content types are mentioned (e.g., ``excessive weight loss + non-suicidal self-injury''), each relevant category must be independently assessed and annotated.
    \item For uncertain or ambiguous mentions, use contextual information to infer whether they truly indicate tendencies or are merely other descriptions (e.g., allusions, metaphors, jokes).
\end{itemize}

\subsubsection{Distinguishing Between Poster vs. Third Party}
\begin{itemize}
    \item \textbf{Tendencies/behaviors occurring to the poster}: If it is clearly evident that the author is discussing or revealing their own behaviors or emotions, mark as ``1''.
    \item \textbf{Only mentioning others' experiences or public events}: If the post primarily describes third parties (friends, family, celebrities, news, etc.), not the author themselves, mark as ``2''.
    \item \textbf{Completely lacking such content}: If no suggestions or comments about DM, ED, or NSSI/SI appear, mark as ``0''.
\end{itemize}

\subsubsection{Concurrent Annotation}
\begin{itemize}
    \item As this employs multi-label annotation, assign values 0/1/2 to each of the three categories (DM, ED, NSSI/SI), resulting in three annotation values.
\end{itemize}

\subsection{Detailed Annotation Guidelines and Examples}

The following examples are intended to illustrate judgment principles; actual annotation should incorporate more contextual information.

\subsubsection{Drug Misuse (DM)}
\begin{itemize}
    \item \textbf{0: No tendency}
    \begin{itemize}
        \item Posts completely lacking mention of drugs, narcotics, misuse, etc.
        \item Example: ``I am in a good mood today, went for a run.''
    \end{itemize}
    \item \textbf{1: Poster's own DM tendency}
    \begin{itemize}
        \item Posts mentioning the author's own drug misuse behaviors, intentions, or plans.
        \item Example: ``I've been relying on taking many sedatives to sleep lately, I wish I could just take enough at once.''
    \end{itemize}
    \item \textbf{2: DM comments from third parties}
    \begin{itemize}
        \item Posts mentioning others' drug misuse or social incidents, not the author's actions.
        \item Example: ``My friend was hospitalized yesterday for taking too many sleeping pills, it's terrifying.''
    \end{itemize}
\end{itemize}

\subsubsection{Eating Disorders (ED)}
\begin{itemize}
    \item \textbf{0: No tendency}
    \begin{itemize}
        \item Posts without any expressions related to eating disorders.
        \item Example: ``I made steak and salad today, it was delicious.''
    \end{itemize}
    \item \textbf{1: Poster's own ED tendency}
    \begin{itemize}
        \item Posts implying or directly stating the author's eating disorder, such as deliberate fasting, purging, excessive dieting, etc.
        \item Example: ``I haven't eaten anything for three days, I feel like I'm about to reach my ideal weight.''
    \end{itemize}
    \item \textbf{2: Third-party ED comments}
    \begin{itemize}
        \item Mentions of others having eating disorder behaviors or discussions of others/social cases.
        \item Example: ``My roommate frequently binges and then purges, I'm a bit worried about her.''
    \end{itemize}
\end{itemize}

\subsubsection{Non-Suicidal Self-Injury and Suicidal Ideation (NSSI/SI)}
\begin{itemize}
    \item \textbf{0: No tendency}
    \begin{itemize}
        \item Completely lacking any hint of non-suicidal self-injury, suicidal thoughts, or behaviors.
        \item Example: ``I made plans with friends to see a movie today, feeling very happy.''
    \end{itemize}
    \item \textbf{1: Poster's own NSSI/SI tendency}
    \begin{itemize}
        \item Clearly mentioning the author's own non-suicidal self-injury or suicidal intentions, behaviors, plans, or strong emotions.
        \item Example: ``I really don't want to live anymore, I've even written a suicide note.''
    \end{itemize}
    \item \textbf{2: Third-party NSSI/SI comments}
    \begin{itemize}
        \item Posts only discussing others' self-injury/suicide situations, social news, seeking help, etc.
        \item Example: ``My classmate was hospitalized for cutting her wrists due to depression, I don't know how to help her.''
    \end{itemize}
\end{itemize}

\subsection{Common Issues and Handling Principles}

\subsubsection{Ambiguous/Vague Information}
\begin{itemize}
    \item If the post doesn't directly mention self or others but has slight implications with uncertain subjects, make reasonable inferences based on context. If unable to determine whether it refers to the author or others, it is recommended to mark as \textbf{1} (leaning toward the author's subtle disclosure) or temporarily retain as \textbf{0} (if insufficient evidence).
    \item If annotation functionality exists, add notes for clarification.
\end{itemize}

\subsubsection{Jokes, Metaphors, or Rhetoric}
\begin{itemize}
    \item If the post appears to be joking or metaphorical, such as ``I can't stop eating sweets, it's practically 'suicidal' sweet intake,'' and clearly not referring to actual suicide or ED, it should typically be marked as \textbf{0}.
    \item Context-based judgment is needed to determine if it's merely exaggerated expression.
\end{itemize}

\subsubsection{Multiple Labels}
\begin{itemize}
    \item A post may simultaneously contain two or three risky health behavior tendencies. For example, if a post discusses both drug misuse and self-injury tendencies, mark ``DM=1, NSSI/SI=1''. If ED is not mentioned, it remains 0.
    \item Evaluate each category's label independently, without mutual influence.
\end{itemize}

\subsubsection{All Three Categories Marked as 0}
\begin{itemize}
    \item This means the post contains no text or implications related to DM, ED, or NSSI/SI, in which case the post should be retained with labels (DM=0, ED=0, NSSI/SI=0).
\end{itemize}

\subsection{Annotation Format}
\begin{enumerate}
        \item Original post text or ID
        \item DM annotation (0/1/2)
        \item ED annotation (0/1/2)
        \item NSSI/SI annotation (0/1/2)
        \item (Optional) Notes field: Brief explanation for ambiguous or controversial annotations.
    \end{enumerate}
\section{Inter-annotator Agreement}
\label{sec:IAA}
\begin{table*}[h]
\centering
\resizebox{\textwidth}{!}{%
\begin{tabular}{lccccc}
\toprule
\hline
\multirow{2}{*}{\textbf{Task}} & \multicolumn{3}{c}{\textbf{Pairwise Cohen's Kappa}} & \multirow{2}{*}{\textbf{Average}} & \multirow{2}{*}{\textbf{Fleiss' Kappa}} \\
\cline{2-4}
 & \textbf{A1 vs. A2} & \textbf{A1 vs. Expert} & \textbf{A2 vs. Expert} & & \\
\hline
Overdose (OD) & 0.7491 & 0.6826 & 0.6434 & 0.6917 & 0.6867 \\
Eating Disorder (ED) & 0.7089 & 0.8498 & 0.7872 & 0.7820 & 0.7844 \\
Non-Suicidal (NSSI) & 0.7122 & 0.8551 & 0.7681 & 0.7785 & 0.7813 \\
\hline
\bottomrule
\end{tabular}
}
\caption{Inter-Annotator Agreement for Content Annotation Tasks}
\label{tab:agreement}
\end{table*}

\section{Prompt}
\label{sec:prompt}

Table \ref{tab:prompts} illustrates the template templates designed to detect risky health behavior in three languages: Chinese, Japanese, and English, and three categories of behavior: DM, ED, and NSSI. These prompts are specifically designed to leverage the zero-shot and two-shot learning capabilities of LLMs, bypassing the need for traditional supervised training with labeled datasets.

Each prompt provides a clear and structured instruction to the LLM, ensuring that the model can classify social media content into one of three predefined categories:

\begin{enumerate}
    \item 0: Content is completely unrelated to the specified behavior.
    \item 1: Content implies or directly mentions the specified behavior.
    \item 2: Content describes or comments on someone else’s behavior within the specified category.
\end{enumerate}

\subsection{Zero-Shot and Two-Shot Use}
These prompts are designed to take advantage of the zero-shot or two-shot capabilities of LLMs:

\begin{enumerate}
    \item Zero-shot settings: The prompts provide all necessary context and instructions within a single input, enabling the LLM to perform the classification task without requiring any prior task-specific examples. For example, in the drug misuse (DM) category, the prompt instructs the LLM to determine whether the content relates to drug abuse and to return a single numerical label as the result.
    \item Two-shot settings: In cases where a slight improvement in performance is desired, the prompts can be augmented with one or two labeled examples before presenting the actual content to classify. These examples illustrate the task expectations and help the LLM align its reasoning more closely with the intended outputs.
\end{enumerate}

\subsection{Multilingual and Multi-Category Design}
Templates are standardized in Chinese, Japanese, and English to ensure consistency in task execution regardless of the language. This multilingual alignment allows the LLM to operate effectively across diverse linguistic contexts without retraining or fine-tuning. Similarly, the three categories (DM, ED, and NSSI) are designed with uniform label definitions to ensure interpretability and comparability across behaviors.

\begin{table*}[t]
\centering
\footnotesize
\begin{tabular}{p{0.05\textwidth}p{0.28\textwidth}p{0.28\textwidth}p{0.28\textwidth}}
\toprule
\textbf{Type} & \textbf{Chinese} & \textbf{Japanese} & \textbf{English} \\
\midrule
DM & \CJKfamily{gbsn}请判断以下社交媒体内容是否涉及药物滥用。请只输出一个数字作为结果：\newline\newline 0：完全不相关\newline 1：暗示或直接提到药物滥用\newline 2：对他人药物滥用行为的描述或评论 & \CJKfamily{min}以下のソーシャルメディアの内容が薬物乱用に関連しているかどうかを判断してください。結果として数字のみを出力してください：\newline\newline 0：全く関連なし\newline 1：薬物乱用を暗示または直接言及\newline 2：他人の薬物乱用行為の描写またはコメント & Determine if the following social media content is related to drug misuse. Please output only one number as the result:\newline\newline 0: Completely unrelated\newline 1: Implies or directly mentions drug misuse\newline 2: Description or comment on someone else's drug misuse behavior \\
\midrule
ED & \CJKfamily{gbsn}请判断以下社交媒体内容是否涉及进食障碍。请只输出一个数字作为结果：\newline\newline 0：完全不相关\newline 1：暗示或直接提到进食障碍\newline 2：对他人进食障碍行为的描述或评论 & \CJKfamily{min}以下のソーシャルメディアの内容が摂食障害に関連しているかどうかを判断してください。結果として数字のみを出力してください：\newline\newline 0：全く関連なし\newline 1：摂食障害を暗示または直接言及\newline 2：他人の摂食障害行為の描写またはコメント & Determine if the following social media content is related to eating disorders. Please output only one number as the result:\newline\newline 0: Completely unrelated\newline 1: Implies or directly mentions eating disorders\newline 2: Description or comment on someone else's eating disorder behavior \\
\midrule
NSSI & \CJKfamily{gbsn}请判断以下社交媒体内容是否涉及自残或自杀倾向。请只输出一个数字作为结果：\newline\newline 0：完全不相关\newline 1：暗示或直接提到自残/自杀\newline 2：对他人自残/自杀行为的描述或评论 & \CJKfamily{min}以下のソーシャルメディアの内容が自傷行為や自殺傾向に関連しているかどうかを判断してください。結果として数字のみを出力してください：\newline\newline 0：全く関連なし\newline 1：自傷/自殺を暗示または直接言及\newline 2：他人の自傷/自殺行為の描写またはコメント & Determine if the following social media content is related to non-suicidal self-injury or suicidal tendencies. Please output only one number as the result:\newline\newline 0: Completely unrelated\newline 1: Implies or directly mentions non-suicidal self-injury/suicide\newline 2: Description or comment on someone else's non-suicidal self-injury/suicide behavior \\
\bottomrule
\end{tabular}
\caption{Prompt templates used for risky health behavior detection across three languages (Chinese, Japanese, and English) and three categories: drug misuse (DM), eating disorders (ED), and non-suicidal self-injury and suicidal ideation (NSSI/SI).}
\label{tab:prompts}
\end{table*}

\section{JiraiLLM-Qwen}
\label{sec:param}
For our JiraiLLM-qwen model, we employed a conservative fine-tuning approach with carefully selected hyperparameters to balance computational efficiency and performance. The model was trained for 3 epochs on a high-performance computing infrastructure consisting of 4 NVIDIA A6000 GPUs, with a relatively small batch size of 5 supplemented by gradient accumulation steps of 10 (effectively creating a virtual batch size of 50) to optimize memory utilization while maintaining training stability. This hardware configuration provided sufficient computational capacity to efficiently process our dataset of 3,000 Chinese samples while minimizing training time.All other hyperparameters were kept at their default values to maintain consistency with established fine-tuning protocols for the Qwen2.5 architecture.

\end{CJK*}
\end{document}